\documentclass[10pt,twocolumn,letterpaper]{article}

\usepackage{wacv}
\usepackage{times}
\usepackage{epsfig}
\usepackage{graphicx}
\usepackage{amsmath}
\usepackage{amssymb}
\usepackage{cite}
\usepackage{float}
\usepackage{tabularx}
\usepackage{enumitem}
\usepackage{multirow}

\def\wacvPaperID{350} 

\wacvfinalcopy 
\ifwacvfinal
\def\assignedStartPage{9876} 
\fi

\ifwacvfinal
\usepackage[breaklinks=true,bookmarks=false]{hyperref}
\else
\usepackage[pagebackref=true,breaklinks=true,colorlinks,bookmarks=false]{hyperref}
\fi

\ifwacvfinal
\else
\fi

\begin{document}

\title{Visual Speech Enhancement Without A Real Visual Stream}

\author{Sindhu B Hegde\thanks{The authors have contributed equally to the work}\\
IIIT Hyderabad\\
{\tt\small sindhu.hegde@research.iiit.ac.in}
\and
K R Prajwal\footnotemark[1]\\
IIIT Hyderabad\\
{\tt\small prajwal.k@research.iiit.ac.in}

\and
Rudrabha Mukhopadhyay\footnotemark[1]\\
IIIT Hyderabad\\
{\tt\small radrabha.m@research.iiit.ac.in}

\and
Vinay Namboodiri\\
University of Bath\\
{\tt\small vpn22@bath.ac.uk}

\and
C.V. Jawahar\\
IIIT Hyderabad\\
{\tt\small jawahar@iiit.ac.in}
}

\maketitle
\begin{abstract}
In this work, we re-think the task of speech enhancement in unconstrained real-world environments. Current state-of-the-art methods use only the audio stream and are limited in their performance in a wide range of real-world noises. Recent works using lip movements as additional cues improve the quality of generated speech over ``audio-only" methods. But, these methods cannot be used for several applications where the visual stream is unreliable or completely absent. We propose a new paradigm for speech enhancement by exploiting recent breakthroughs in speech-driven lip synthesis. Using one such model as a teacher network, we train a robust student network to produce accurate lip movements that mask away the noise, thus acting as a ``visual noise filter". The intelligibility of the speech enhanced by our pseudo-lip approach is comparable ($< 3\%$ difference) to the case of using real lips. This implies that we can exploit the advantages of using lip movements even in the absence of a real video stream. We rigorously evaluate our model using quantitative metrics as well as human evaluations. Additional ablation studies and a demo video on our website containing qualitative comparisons and results clearly illustrate the effectiveness of our approach. We provide a demo video which clearly illustrates the effectiveness of our proposed approach on our website: \scriptsize{\url{http://cvit.iiit.ac.in/research/projects/cvit-projects/visual-speech-enhancement-without-a-real-visual-stream}}. 
\normalsize{The code and models are also released for future research:} \scriptsize{\url{https://github.com/Sindhu-Hegde/pseudo-visual-speech-denoising}}.
\end{abstract}

\section{Introduction}
\label{section:introduction}

Imagine calling your friend from inside a crowded public bus. She is unable to hear your plans for the evening due to the noise of the bus, wind, and the nearby moving vehicles. We are all constantly surrounded by noise that corrupts our speech. The problem of speech enhancement is thus quintessential, especially at a time when several work-related meetings are happening over a phone call from our homes. But the applications of speech enhancement extend well beyond voice calls. For instance, separating human speech from the background music can be crucial for automatic subtitle/lyrics generation for movies and music. Further, speech enhancement can help the rising number of independent content creators filter the outdoor noises that are widely prevalent in their vlogs and short videos. Last but not least, enhancing historically important speeches will help us preserve our heritage for future generations.

\subsection{Overview of Existing Approaches}
The problem of speech enhancement has been studied for a long time, and various methods have already been proposed. Recently, several works~\cite{8461944,segan,Germain2019SpeechDW} using deep learning have become popular, where noisy speech is enhanced using only the audio modality. However, all these works are applicable only in mild noise conditions (high SNR) and are known to produce artifacts in the generated speech. Most importantly, they often do not produce satisfactory results for unconstrained real-world applications such as those mentioned above. This is because various types of unseen noises degrade the input audio, the speakers and the recording systems go through unforeseen changes. A newer trend was introduced by works~\cite{cocktailparty_erphat_2018,TheConversation_Afouras_2018} where lip movements are used as an additional modality for enhancement. These methods are more accurate than audio-only works in unconstrained settings with large amounts of noise levels. However, unlike audio-only works, audio-visual methods are adversely affected by rapid head motion, occlusions, lip going out of focus, or the audio and lips going out of sync~\cite{TheConversation_Afouras_2018}. This significantly limits the applicability of these audio-visual methods, where most of the common applications discussed above cannot be handled due to the absence of a visual stream with clearly visible lip movements.

\begin{figure*}[t]
  \includegraphics[width=\textwidth]{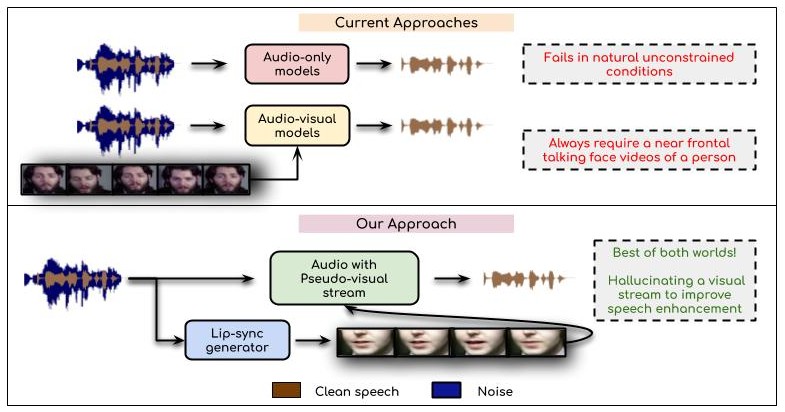}
  \caption{We propose a novel approach to enhance the speech by hallucinating the visual stream for any given noisy audio. In contrast to the existing audio-visual methods, our approach works even in the absence of a reliable visual stream, while also performing better than audio-only works in unconstrained conditions due to the assistance of generated lip movements.}
  \label{fig:firstpage}
\end{figure*}

\subsection{A Visual Noise Filter}
In this work, we introduce a new hybrid paradigm that brings together the best of both these branches of speech enhancement. We want a method that exploits the robustness and accuracy boost one can get with lip movements but also be able to function effectively in a wide range of applications where the visual stream is unreliable or absent. Thus, instead of relying on an actual video stream, we propose to use a lip synthesis model to generate a visual stream with lip movements for any given noisy audio. Using a carefully designed student-teacher training setup, we demonstrate that we can generate face images with lip movements that do not meaningfully represent the noise but accurately reflect the underlying speech component. Consequently, the generated images act as a \textit{visual noise filter} for the down-stream speech enhancement model. In this way, unlike the existing audio-visual enhancement works, we are not constrained by the need for a video with clear visibility of lip movements. However, we can still utilize the improvements these works achieve with the help of pseudo lip movements (Figure \ref{fig:firstpage}). To the best of our knowledge, this is the first work to grasp the advantages of the visual stream, even in the presence of only audio. Our proposed approach yields significant improvements across all speech quality and intelligibility metrics and human evaluation studies. To summarize, the following are our major contributions:

\begin{itemize}[nosep]
     \item We propose a novel pseudo-visual speech enhancement model that is applicable in natural and high noise conditions.  
     
     \item We are the first to study the use of artificial lip movements from a lip synthesis model for speech enhancement. Our method is the first to effectively use the benefits of lip movements and still be applicable in situations where the visual information is unavailable or is corrupted.
     
     \item Using a novel student-teacher setup, we show that we can train a lip synthesis model to generate accurate lip movements corresponding to the underlying speech in a noisy signal. In fact, the intelligibility of the speech enhanced by our pseudo-lip approach is comparable ($< 3\%$ difference) to the case of using real lips.
     
     \item We create and release a new standard human evaluation set consisting of real-world videos in unconstrained conditions with several types of noises. Future speech enhancement works can evaluate their perceptual quality on this set. 
     
\end{itemize}

We provide a demo video on our website, which clearly exhibits our approach's feasibility compared to the existing audio-only and audio-visual works. The code, trained models, and the evaluation benchmark are released publicly for future research\footnote{\scriptsize\url{cvit.iiit.ac.in/research/projects/cvit-projects/visual-speech-enhancement-without-a-real-visual-stream}}. The rest of the paper is organized as follows: In Section~\ref{section:related}, we give an overview of the existing works in this space. We then elaborate on the proposed speech enhancement method in Section~\ref{section:methodology}. The experimental setup, results and analysis are discussed in Sections~\ref{section:experiments} and~\ref{section:results}. We perform several ablation studies in Section~\ref{section:ablation} and discuss various potential applications of our work in Section~\ref{section:applications}. We conclude our work in Section~\ref{section:conclusion}. 

\section{Related Work}
\label{section:related}
\subsection{Audio-only Speech Enhancement}
We first review the works using only the audio stream with no additional information for denoising of speech. Classical signal denoising techniques like the Wiener filtering~\cite{wienerfilter} became the first popular approach for speech enhancement. However, it was often ineffective in denoising speech in real-world situations as the wiener filter requires an estimate of the noise a priori. Like many other problems, deep learning models have become increasingly popular for speech enhancement in recent times. Initial works~\cite{boltzmann_2015_sui,Lu2013SpeechEB} used standard denoising auto-encoders and LSTM based approaches~\cite{8461944, ValentiniBotinhao2016InvestigatingRS} for cleaning noisy speech. Feature-based loss functions have also been proposed in~\cite{Germain2019SpeechDW}, while the most popular advancement came from models like~\cite{segan, Donahue2018ExploringSE,Germain2019SpeechDW,ConditionalGA_2017_Michelsanti} using generative adversarial networks (GANs) which produce relatively higher quality speech from noisy audio segments.  

Even though there has been significant progress in the last few years, speech enhancement models are still confined to being trained on datasets~\cite{ValentiniBotinhao2016InvestigatingRS, Dean2010TheQC,6709856, timit} that has been collected in constrained environments recorded by a selected set of speakers. The types of noises~\cite{ValentiniBotinhao2017NoisySD, piczak2015dataset} that are synthetically added to the clean speech while training such models are also limited to a few types. Thus, these models often do not perform well in unconstrained natural settings as they fail to cope up with hundreds of speakers of different dialects and languages, the level and type of noise varying abruptly in a speech segment, etc. In this work, we aim to generate high-quality clean speech from the given noisy audio in such unconstrained conditions.  

\subsection{Audio-visual Speech Enhancement}
Since 2018, a new approach was introduced for denoising of speech. Works like~\cite{cocktailparty_erphat_2018,TheConversation_Afouras_2018, Owens_2018_ECCV} considered an additional visual stream of information by exploiting the lip movements of the speaker for extracting the clean speech. These methods not only perform well in unconstrained real-world settings, but they also show significant improvement in terms of metrics over other audio-only methods. However, they suffer from a major limitation that they work only on videos with a clear view of the speaker's lips. This requirement of a frontal, lip-synced video of the speaker prohibits these models from being used for a wide range of applications, where the visual stream is imperfect (profile views, lips going out of focus, video corruption, out-of-sync speech, motion blurs) or even completely absent.

\subsection{Understanding Speech and Lip Movements}
Jointly understanding multiple modalities together has gained significant traction in recent times. Several recently published works~\cite{Perez_2020_WACV,Wang_2020_WACV,Gao_2018_CVPR_Workshops} involve both audio and vision as modalities. Interpreting the spoken utterances from lip movements has attracted special attention in different works~\cite{Prajwal_2020_CVPR,Afouras18c,Afouras18d,lr1_petridis_2017,lr2_petridis_2017,lr3_petridis_2020}. The opposite task of generating lip motion for given speech segments has also become a popular area of research. The initial works in this space~\cite{Suwajanakorn2017SynthesizingOL,obamanet} were specifically trained for a single person with several hours of data. More recent works~\cite{wav2lip:2020, KR:2019:TAF:3343031.3351066,yst_ijcv_2019} can generate lip movements for any identity, any voice, and any language. Specifically, Wav2Lip~\cite{wav2lip:2020} is the current state-of-the-art in ``unconstrained lip syncing" which produces accurate lip motion for any given speech, but is inaccurate for noisy speech. In the next section, we discuss in detail how we can distill the knowledge of this network into a student network that learns to generate accurate lip movements for a given noisy speech segment.

\section{Pseudo-visual Stream for Enhancing Speech}
\label{section:methodology}
How do we generate lip motion that is in sync with the clean speech component in given noisy audio?

\paragraph{Can we readily use the current lip synthesis models for noisy speech inputs?} We find that the current state-of-the-art unconstrained speech-to-lip models~\cite{wav2lip:2020, KR:2019:TAF:3343031.3351066,yst_ijcv_2019} which work for arbitrary speakers, voices, and languages are highly inaccurate on noisy speech segments. This is understandable as these works were never aimed to tackle such cases. Moreover, these methods are speaker-independent and are designed to work on unconstrained videos by training on thousands of speakers with substantial variations in pose, expressions, backgrounds, etc. We found that it is not ideal to naively fine-tune the pre-trained lip synthesis model on noisy speech. This is simply because learning to lip-sync on highly noisy speech across such extreme variations in the visual data is a daunting task, yielding limited improvements in fine-grained lip shapes (Table \ref{table:lipsync_metric}). But, for our task at hand, we do not need a lip synthesis model on thousands of speakers; a single speaker is sufficient.

\begin{figure}[t]
  \includegraphics[width=\linewidth]{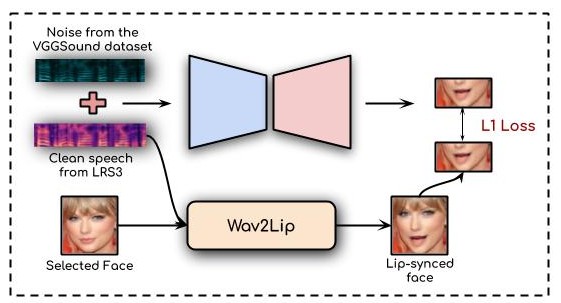}
  \caption{We train a novel student-teacher network for generating accurate lip movements for noisy speech segments. The teacher is a pre-trained lip synthesis network~\cite{wav2lip:2020} that generates accurate lip movements on a static face using clean speech as input. The student is trained to mimic the teacher's lip movements, but when given noisy speech as input.}
  \label{fig:lip_synthesis}
\end{figure}

\begin{figure*}[t]
  \includegraphics[width=\linewidth]{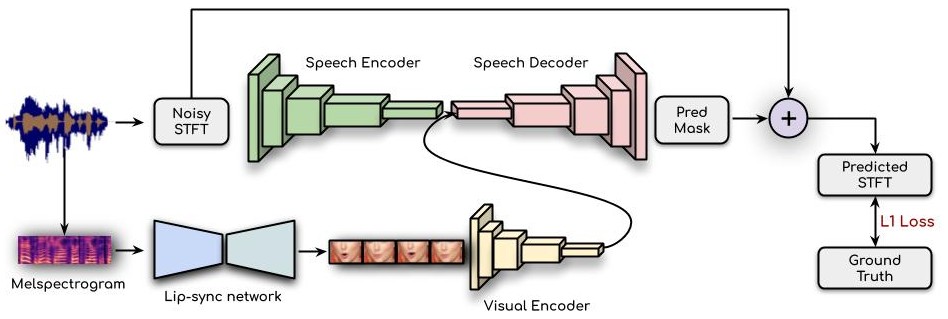}
  \caption{Our proposed speech enhancement model. A pseudo-visual stream is generated for the given noisy audio, which acts as a visual noise filter. The enhancement model then ingests the noisy  spectrogram along with the generated lip movements and outputs a mask for the clean speech.}
  \label{fig:av_network}
\end{figure*}

\vspace{-10pt}
\paragraph{A single identity is all you need.}
We only need a sequence of accurate lip movements, preferably even just on a static image of a single identity where only the lips are moving in accordance with the speech. This is a relatively much easier task than the one before and is also well-aligned with our needs. If the only visual changes are in the lip shapes, the model will naturally focus on learning more accurate, fine-grained speech-lip correspondences. Note that we still need this identity-specific model to work for any speech in any voice and language. How do we train a lip synthesis model that can lip-sync just for a single identity image, but can handle any speech?

\vspace{-10pt}
\paragraph{Distilling lip motion knowledge for a single identity}
We exploit the fact that the current state-of-the-art model, Wav2Lip~\cite{wav2lip:2020} can generate accurate lip motion for arbitrary static face images conditioned on any \textit{clean} speech. Our core idea is to achieve this accuracy using noisy speech (harder part) as input, but on just a single identity (easier part). To do this, we train a student network to map the noisy speech inputs to lip motion on a single static face image. We employ Wav2Lip as the teacher network and use its predictions on the same identity image, but with clean speech inputs. This is illustrated in Figure \ref{fig:lip_synthesis}. 

\vspace{-10pt}
\paragraph{A Visual Noise Filter:} We hypothesize that since the only visual differences are in the lip shapes, the student network is forced to learn a strong correspondence between the underlying speech and the lip motion. Further, the student network cannot meaningfully represent noise in the generated images and is forced to represent only the speech components that the teacher network accurately indicates. Thus the images generated by the student network acts as a ``visual noise filter" that manifests only the speech component for the down-stream speech enhancement network.

\subsection{Training the Student Model}
\label{subsection:step2}
As described above, we would like to train a student model $M$ by learning from a pre-trained lip-synthesis network $L$ as a teacher. $M$ is a simple encoder-decoder model that inputs a noisy speech segment and outputs a lip-synced mouth region of a pre-determined person. This is adapted from the Wav2Lip architecture~\cite{wav2lip:2020} by discarding the face identity branch, because we need $M$ to generate lip movements only for a single identity image.

\paragraph{Noisy Speech Input to $M$:} We feed a $0.2$ second window of the noisy speech segment. We create this noisy speech segment ($S_{input}$) by mixing clean speech ($S_{clean}$) from the LRS3~\cite{Afouras18d} with noise ($S_{noise}$) from the VGGSound~\cite{9053174} dataset at one of the three signal-to-noise ratios (SNR) ($0$, $5$ and $10$ dB). As done in Wav2Lip~\cite{wav2lip:2020}, we use mel-spectrograms as the input speech representation.

\paragraph{Learning from a Lip Synthesis Teacher}
To train the model $M$, we need an accurate lip-synced ground-truth of a single target. We obtain this from $L$, a pre-trained lip-synced network, by feeding the clean speech $S_{clean}$ as the audio input. We use the audios present in the LRS3~\cite{Afouras18d} dataset as our clean speech data. For our lip synthesis teacher $L$, we use Wav2Lip~\cite{wav2lip:2020}, a publicly available\footnote{\url{https://github.com/Rudrabha/Wav2Lip}} state-of-the-art speech-to-lip synthesis model. As it is a speaker-independent model, we also need to feed an identity image. We choose a near-frontal face image of Taylor Swift on which the lips are morphed by Wav2Lip to match the clean speech inputs. The lip-synced output from Wav2Lip is accurate as the audio is clean. Further, the output is always of the same face image with only the lip and jaw regions changing while the rest of the face regions remain static. We use the lower half of the generated face output containing the Wav2Lip's prediction as ground-truth for our student model $M$. Thus, our new student network is trained to generate correct lip movements (matching the clean speech) given a noise-corrupted input of the same speech. 

We train the student network to minimize the $L1$ loss between its predicted images and the lip-synced ground truth from Wav2Lip. We train this network for $150K$ iterations with a batch size of $64$ on a single NVIDIA RTX 2080Ti GPU. Other hyper-parameters are the same as that of Wav2Lip~\cite{wav2lip:2020}. For our speech enhancement model in the next section, we use this trained student network to generate the lip motion given a noisy speech segment.

\subsection{Pseudo-visual Speech Enhancement}
\label{subsection:step3}

An overview of the proposed speech enhancement model is illustrated in Figure~\ref{fig:av_network}. Our model takes both the pseudo-visual stream and the noisy auditory stream as the input. For a given noisy audio, initially, we generate the lip movements as described in Section~\ref{subsection:step2}. These, along with the noisy input spectrograms are given to the visual encoder, and the speech encoder respectively as shown in Figure~\ref{fig:av_network}. The speech decoder outputs a residual mask, which is added to the input spectrograms to filter the noise from the clean speech.

\subsubsection{Audio representation:} We consider $1$ second of noisy speech, $S_{input}$ as input to the enhancement model and extract linear spectrogram representation using short-time Fourier
transform (STFT). Generating linear spectrograms allows us to directly invert them back to a waveform without the need for vocoders. To compute the STFT, we consider the window length of $25$ms with a hop length of $10$ms sampled at $16$kHz. The computed STFT from the raw audio waveforms is a complex array of time-frequency representation, with a dimension of $T_s \times 257$. Here, $T_s$ is the number of STFT time steps, which corresponds to $100$ in our experiments ($1$ second audio segment). We further decompose the complex STFT array into the magnitude and the phase components, and normalize them between $[0, 1]$. These components are concatenated along the frequency axis to form a representation of $T_s \times 514$ which acts as input to the speech encoder network. 

\vspace{-10pt}
\subsubsection{Visual representation:} The lip-sync student network generates $25$ frames for one second of audio input. Using a visual encoder consisting of $12$ layers of residual 2D-convolution blocks, we obtain a visual embedding for each frame. 

\vspace{-10pt}
\subsubsection{Network architecture:} The speech enhancement model consists of the speech encoder and decoder networks along with a visual encoder to encode the pseudo-visual stream as illustrated in Figure~\ref{fig:av_network}. 
The input noisy spectrogram is processed by the speech encoder which is a stack of $7$ 1D-convolution blocks with residual connections. We perform the convolutions along the temporal dimension, by considering the frequency component of the input spectrograms as channels. The output of the visual encoder module is up-sampled $4\times$ using nearest-neighbor interpolation to match the spectrogram temporal dimension. We then combine the audio and the visual streams by concatenating the learned features of each stream along the channel dimension. This fused representation is then given to the speech decoder which is a stack of $14$ 1D convolution layers with residual connections. The decoder outputs a mask that is added to the input noisy spectrogram followed by a sigmoid activation to generate the enhanced speech spectrogram output. We minimize the $L1$ distance between the predicted spectrogram and the ground truth. Finally, the enhanced clean waveform is obtained by using inverse-STFT (ISTFT).

\section{Experiments}
\label{section:experiments}

\subsection{Dataset}
We use the publicly available LRS3 dataset~\cite{Afouras18d} which consists of thousands of spoken sentences from TED videos. For training, we use the ``pre-train" and ``train-val" sets from the dataset which has around $430$ hours of video data with $150K$ utterances. This is a challenging dataset that covers a large number of speakers ($9K$), thus encouraging the trained model to be speaker-independent.

\subsection{Experimental setup}
\label{setup}
For evaluating in unconstrained settings, we create the following three synthetic test sets, each having three noise levels of 0db, 5db and 10db: (i) Test split of LRS3~\cite{Afouras18d} + Noise from VGGSound~\cite{9053174}, (ii) Test split of LRS3~\cite{Afouras18d} + Noise from QUT-NOISE-TIMIT~\cite{Dean2010TheQC} (unseen noise), (iii) Test split of LRS2~\cite{Afouras18c} + Noise from VGGSound~\cite{9053174} (unseen speakers). 

\subsection{Evaluation}
\label{subsection:test_sets}
We use the following standard speech enhancement evaluation metrics (higher is better) to evaluate our method. We compute Perceptual Evaluation of Speech Quality (PESQ)~\cite{rix2001perceptual} ($-0.5$ to $4.5$), which measures the overall perceptual quality and short-time objective intelligibility measure (STOI)~\cite{taal2010short} ($0$ to $1$), which correlates with the intelligibility of speech. We also use objective measures such as the mean opinion score (MOS) prediction of the signal distortion (CSIG) ($1$ to $5$), the MOS prediction of background noise (CBAK) ($1$ to $5$), and the overall MOS prediction score (COVL) ($1$ to $5$). In addition to evaluation based on these metrics, we also perform human evaluations on a newly curated real-world test set and report the MOS (range: $1$ to $5$) to analyze the real-world applicability of our model.

\section{Results}
\label{section:results}

\subsection{Evaluating the Lip-sync Network}
\label{subsection:lipsync_eval}

We start by evaluating our lip-sync network that generates lip movements for a given noisy speech segment. It should be noted that this network is specifically designed for our speech enhancement task and not for the purpose of talking face generation. For comparison, we also fine-tune Wav2Lip~\cite{wav2lip:2020} with the same noisy speech samples used for training the student network. For evaluation, we use the synthetic test set consisting of speech from LRS3 data and noise from VGGSound as described in Section~\ref{setup}.

We then use this test set to benchmark (a) pre-trained Wav2Lip, (b) Wav2Lip trained on noisy data, and (c) Our student lip-sync network. Note that there are no original ground truth videos available for the kind of talking face videos our lip-sync model generates from a single face image. Thus, we use the LSE-D and LSE-C metrics used to evaluate lip-sync in Wav2Lip~\cite{wav2lip:2020}. A higher LSE-C indicates better overall audio-visual correlation. As we can see from Table~\ref{table:lipsync_metric}, our approach to train a lip-sync network specifically for this task with one face using Wav2Lip as a teacher outperforms other methods in terms of LSE-C, indicating a better overall audio-visual correspondence. The LSE-D of the student is close to the pre-trained Wav2Lip, but the confidence score, LSE-C of the pre-trained Wav2Lip is quite low. In Section~\ref{subsection:diff_visual_stream}, we also show the performance of our speech enhancement model when using other networks to generate the pseudo-visual stream. We now move on to evaluating our main pipeline for speech enhancement. 

\begin{table}[ht]
    \centering
    \caption{Quantitative Evaluation of the lip sync model. Directly using the pre-trained Wav2Lip model or fine-tuning it on noisy audios leads to poor results on noisy speech. Our student network leads to more clear audio-visual correspondence as denoted by LSE-C.}
    \begin{tabular}{lccc}
    \hline
    \textbf{Model} & LSE-C$\uparrow$ & LSE-D$\downarrow$\\
    \hline
    Wav2Lip trained on noisy data & 1.181 &  10.35\\
    Pre-trained Wav2Lip & 1.330 & \textbf{8.933} \\
    \textbf{Lip-sync student (ours)} & \textbf{4.572} & 9.743\\
    
    \hline
    \end{tabular}
    \label{table:lipsync_metric}
\end{table}

\subsection{Quantitative Evaluation}

\begin{table*}[ht]
  \setlength{\tabcolsep}{3.1pt}
    \centering
    \caption{Quantitative comparison of different approaches. The first section contains clean speech from LRS3~\cite{Afouras18d} test set mixed with VGGSound~\cite{9053174} noises at different SNR levels. In the second section, we specifically evaluate the performance on ``unseen noises" by mixing the LRS3~\cite{Afouras18d} test set audios with the QUT~\cite{9053174} city-street noises at different noise levels. Finally, in the third section, we evaluate specifically on ``unseen speakers" by mixing the speeches of the unseen LRS2~\cite{Afouras18c} test set speakers with VGGSound~\cite{9053174} noises. Our method outperforms the audio-only approaches in all three sections and is comparable ($< 3\%$ difference) to the real visual-stream method.}
    \begin{tabular}{l||ccc|cc|c||ccc|cc|c||ccc|cc|c}
    \hline
    \textbf{SNR} & \multicolumn{6}{c||}{0db} & \multicolumn{6}{c||}{5db} & \multicolumn{6}{c}{10db}
     \\
    \hline
    \textbf{Method} & 
    Noisy & \cite{segan} & \cite{Germain2019SpeechDW} & AO & Ours & \cite{TheConversation_Afouras_2018} & 
    Noisy & \cite{segan} & \cite{Germain2019SpeechDW} & AO & Ours & \cite{TheConversation_Afouras_2018} &  
    Noisy & \cite{segan} & \cite{Germain2019SpeechDW} & AO & Ours & \cite{TheConversation_Afouras_2018} \\
    \hline
    \textbf{PESQ} & 1.93 & 1.84 & 2.17 & 2.62 & \textbf{2.72} & 2.80 &
                    2.29 & 2.24 & 2.52 & 2.93 & \textbf{2.99} & 3.05 &
                    2.66 & 2.65 & 2.95 & 3.12 & \textbf{3.19} & 3.25\\
    \textbf{CSIG} & 2.31 & 2.10 & 2.82 & 3.02 & \textbf{3.18} & 3.25 &
                    2.79 & 2.67 & 3.22 & 3.26 & \textbf{3.32} & 3.39 &
                    3.15 & 3.17 & 3.36 & 3.45 & \textbf{3.51} & 3.56\\
    \textbf{CBAK} & 1.83 & 1.87 & 2.30 & 2.36 & \textbf{2.47} & 2.51 &
                    2.23 & 2.30 & 2.46 & 2.54 & \textbf{2.65} & 2.71 &
                    2.40 & 2.55 & 2.63 & 2.70 & \textbf{2.81} & 2.84\\
    \textbf{COVL} & 1.68 & 1.53 & 2.02 & 2.14 & \textbf{2.25} & 2.29 &
                    2.04 & 2.00 & 2.21 & 2.29 & \textbf{2.37} & 2.41 &
                    2.16 & 2.26 & 2.31 & 2.46 & \textbf{2.52} & 2.58\\
    \textbf{STOI} & 0.76 & 0.75 & 0.83 & 0.87 & \textbf{0.88} & 0.90 &
                    0.85 & 0.86 & 0.88 & 0.90 & \textbf{0.92} & 0.94 &
                    0.88 & 0.88 & 0.90 & 0.92 & \textbf{0.95} & 0.95\\
    \hline
    \hline
    \textbf{PESQ} & 1.86 & 1.77 & 2.14 & 2.54 & \textbf{2.65} & 2.73 &
                    2.26 & 2.14 & 2.54 & 2.83 & \textbf{2.92} & 3.01 &
                    2.67 & 2.61 & 2.90 & 3.05 & \textbf{3.12} & 3.21\\
    \textbf{CSIG} & 2.46 & 2.37 & 2.78 & 2.93 & \textbf{3.02} & 3.10 &
                    2.90 & 2.86 & 3.18 & 3.15 & \textbf{3.24} & 3.33 &
                    3.30 & 3.30 & 3.38 & 3.35 & \textbf{3.42} & 3.48\\
    \textbf{CBAK} & 1.55 & 1.89 & 2.10 & 2.31 & \textbf{2.43} & 2.46 &
                    1.94 & 2.23 & 2.30 & 2.49 & \textbf{2.59} & 2.63 &
                    2.42 & 2.53 & 2.59 & 2.62 & \textbf{2.73} & 2.77\\
    \textbf{COVL} & 1.69 & 1.68 & 1.95 & 2.04 & \textbf{2.12} & 2.20 &
                    2.02 & 1.99 & 2.06 & 2.20 & \textbf{2.29} & 2.34 &
                    2.14 & 2.20 & 2.30 & 2.35 & \textbf{2.41} & 2.45\\
    \textbf{STOI} & 0.75 & 0.77 & 0.80 & 0.84 & \textbf{0.87} & 0.89 &
                    0.85 & 0.86 & 0.87 & 0.89 & \textbf{0.91} & 0.93 &
                    0.87 & 0.90 & 0.91 & 0.91 & \textbf{0.94} & 0.95\\
    \hline
    \hline
    \textbf{PESQ} & 1.94 & 1.82 & 2.10 & 2.58 & \textbf{2.71} & 2.79 &
                    2.32 & 1.87 & 2.55 & 2.87 & \textbf{2.98} & 3.04 &
                    2.69 & 2.41 & 2.79 & 3.10 & \textbf{3.19} & 3.22\\
    \textbf{CSIG} & 2.55 & 2.29 & 2.80 & 3.07 & \textbf{3.16} & 3.23 &
                    3.01 & 2.85 & 3.16 & 3.21 & \textbf{3.32} & 3.38 &
                    3.23 & 3.13 & 3.33 & 3.36 & \textbf{3.44} & 3.49\\
    \textbf{CBAK} & 1.86 & 1.82 & 2.22 & 2.31 & \textbf{2.41} & 2.47 &
                    2.28 & 2.12 & 2.42 & 2.48 & \textbf{2.57} & 2.63 &
                    2.58 & 2.59 & 2.57 & 2.63 & \textbf{2.71} & 2.73\\
    \textbf{COVL} & 1.81 & 1.59 & 1.93 & 2.07 & \textbf{2.15} & 2.20 &
                    2.16 & 2.03 & 2.14 & 2.17 & \textbf{2.26} & 2.31 &
                    2.20 & 2.11 & 2.23 & 2.27 & \textbf{2.35} & 2.39\\
    \textbf{STOI} & 0.75 & 0.73 & 0.80 & 0.84 & \textbf{0.87} & 0.89 &
                    0.84 & 0.83 & 0.85 & 0.88 & \textbf{0.90} & 0.90 &
                    0.85 & 0.86 & 0.89 & 0.89 & \textbf{0.92} & 0.93\\
    \hline
    \end{tabular}
    \label{quant}
\end{table*}

We start by evaluating our method using the synthetic test sets created as described in Section~\ref{setup}.  We present the metric scores of the input noisy data (without any denoising and enhancement) along with other speech enhancement methods. For fair comparison, we fine-tune the audio-only models, SEGAN~\cite{segan} and DFL~\cite{Germain2019SpeechDW} on the same LRS3 training data. To further highlight the importance of our pseudo-visual stream, we implemented our own audio-only (AO) baseline, which is very similar to our model but without the visual encoder. Also, we compare our results with our implementation of the real-visual stream state-of-the-art audio-visual method~\cite{TheConversation_Afouras_2018} to see how close our pseudo-visual approach is to the real visual-stream method. 

\subsubsection{Our model is robust to various noise levels}
The results on LRS3 test set mixed with noise from VGGSound data at different noise levels is summarized in the first section of Table~\ref{quant}. As we can observe, our pseudo-visual model outperforms the audio-only approaches as well as our AO-baseline by a significant margin at all three noise levels. Also, it is very close to the real-visual stream method, which indicates that our model is effective in generating accurate lip movements. It is interesting to note that our model performs comparable to the real-visual stream approach even at higher noise level of 0db. This validates our claim that the generated pseudo-visual stream reflects the clean speech segment and thus is able to suppress the noise. Moreover, another important point to note is that our AO-baseline performs comparably to the existing audio-only methods. 

\subsubsection{Our model is robust to unseen noises}
To further illustrate the generalization of our approach, we present the results on new, unseen noise type from QUT (city-street noise) dataset~\cite{Dean2010TheQC} in the second section of Table~\ref{quant}. We can observe the robustness and the superior performance of our model even on unseen noise types. 

\subsubsection{Our model is robust to unseen speakers}
\begin{figure}[ht]
  \includegraphics[width=\linewidth]{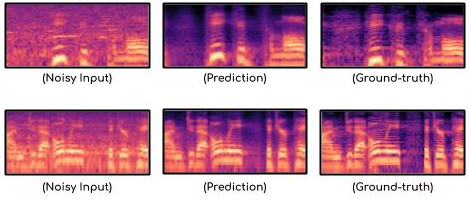}
  \caption{We can clearly see that our network is able to reconstruct clean speech (center spectrogram) which is very close to the ground-truth (right).}
  \label{fig:speech_res}
  \vspace{-10pt}
\end{figure}

We also perform an additional comparative study on unseen speakers from LRS2 test set as shown in the third section of Table~\ref{quant}. In-line with the previous results on LRS3 test set, our method performs remarkably well in comparison with the audio-only approaches. The results clearly indicate that our method is robust to unseen speakers, which is also validated using our collected real-world test set. A sample spectrogram predicted by our model, along with the ground truth and the noisy input spectrograms are shown in Figure~\ref{fig:speech_res}. We observe that our model is able to reconstruct accurate speech even from a highly noisy input.

\subsection{Human Evaluations}
To evaluate the effectiveness of our work in real-world situations, we perform a rigorous human evaluation on a newly collected real-world evaluation set. This set comprises $50$ real-world videos in unconstrained environments, which are originally degraded by different kinds of noises. The test set contains a wide variety of videos/audios such as people vlogging while riding a motorbike or sailing in rough oceans, interaction on camera in crowded airports and railway stations, and old heritage recordings. To the best of our knowledge, such an evaluation set is the first of its kind and can be used for perceptually evaluating the performance of future works on real-world examples. 

As these samples are naturally corrupted by noise, and we do not have the ground truth clean speech, we depend on human evaluations for this dataset. For comparison, we also conduct a human evaluation on the subset containing $25$ samples each from LRS2 and LRS3 datasets when mixed with noise from the VGGSound corpus. In Table~\ref{mos}, we report the mean scores of $15$ participants for different approaches on a scale of $1$-$5$ based on: (A) \textit{Quality}, and (B) \textit{Intelligibility}. The participant group consists of almost equal male-female members spanning an age group of 22 - 49 years. Table~\ref{mos} shows that the speech generated by our model is preferred over the other methods, even in real-world conditions.

\begin{table}[ht]
    \centering
    \setlength{\tabcolsep}{3pt}
    \caption{Human evaluation of our model based on: (A) Quality and (B) Intelligibility.}
    \begin{tabular}{lcccccc}
    \hline
    Data & Metric & Noisy & ~\cite{segan} & ~\cite{Germain2019SpeechDW} & AO & Ours  \\
    \hline
    
    \multirow{2}{*}{LRS2} & (A) & 1.23 & 2.02 & 2.38 & 3.45 & \textbf{3.73}\\
    & (B) & 2.81 & 3.05 & 3.27 & 3.96 & \textbf{4.12}\\
    
    \hline
    
    \multirow{2}{*}{LRS3} & (A) & 1.25 & 2.14 & 2.49 & 3.52 & \textbf{3.84} \\
    & (B) & 2.85 & 3.12 & 3.36 & 3.98 & \textbf{4.25} \\
    
    \hline
    
    Curated real & (A) & 1.49 & 1.92 & 2.27 & 3.31 & \textbf{3.83} \\
    world samples & (B) & 2.66 & 2.91 & 3.08 & 4.01 & \textbf{4.31} \\
    
    \hline
    \end{tabular}
    \label{mos}
\end{table}

\section{Ablation Studies}
\label{section:ablation}

In order to understand our design choices in our approach, we conduct ablation experiments. Unless specified, all the results are reported on the test set of LRS3~\cite{Afouras18d} dataset when mixed with VGGSound~\cite{9053174} at a noise level of 0db. 

\subsection{Model is invariant to the pseudo-lip identity}

In Table~\ref{table:multiple_identities}, we report the performance when different identities are used for the generation of lip movements. Our model is consistent across multiple pseudo-lip identities varied based on gender, age, ethnicity, etc.  

\begin{table}[ht]
    \centering
    \setlength{\tabcolsep}{3pt}
    \caption{The performance of our model is consistent if the age/ethnicity of the pseudo-lip identity is varied.}
    \begin{tabular}{lccccc}
    \hline
 
    \textbf{Identity} & \textbf{PESQ} & \textbf{CSIG} & \textbf{CBAK} & \textbf{COVL} & \textbf{STOI}\\
    \hline
    Taylor Swift & 2.72 & 3.18 & 2.47 & 2.25 & 0.88\\
    Paul McCartney  & 2.70 & 3.19 & 2.45 & 2.24 & 0.88\\
    Morgan Freeman & 2.71 & 3.16 & 2.42 & 2.21 & 0.87\\
    Andrew Ng & 2.72 & 3.18 & 2.46 & 2.23 & 0.88\\

    \hline
    
    \end{tabular}
    \label{table:multiple_identities}
\end{table}

\vspace{-10pt}
\subsection{Comparison of visual stream generators}
\label{subsection:diff_visual_stream}

Although Wav2Lip~\cite{wav2lip:2020} is the state-of-the-art in ``unconstrained lip-syncing", we also compare other lip-sync models for the generation of pseudo-visual stream in Table~\ref{table:lip_generation}. All the methods other than our trained student lip-sync model result in an inferior performance due to the inaccurate lip-shape generation. The main reason is that other methods are constrained and/or sensitive to noise in the speech, and generate wide open mouth shapes in the presence of noise.

\begin{table}[ht]
    \centering
    \setlength{\tabcolsep}{3pt}
    \caption{Comparison of different methods for generation of pseudo-visual stream. Our student lipsync network generates accurate lip movements, thus effectively enhances the noisy speech.}
    \begin{tabular}{lccccc}
    \hline
    \textbf{Lipsync models} & \textbf{PESQ} & \textbf{CSIG} & \textbf{CBAK} & \textbf{COVL} & \textbf{STOI}\\
    \hline
    Speech2Vid~\cite{yst_ijcv_2019} & 2.42 & 2.85 & 2.17 & 2.01 & 0.82   \\
    LipGAN~\cite{KR:2019:TAF:3343031.3351066} & 2.54 & 2.98 & 2.21 & 2.05 & 0.85\\
    Wav2Lip~\cite{wav2lip:2020} & 2.64 & 3.05 & 2.28 & 2.12 & 0.86\\
    \textbf{Ours} & \textbf{2.72} & \textbf{3.18} & \textbf{2.47} & \textbf{2.25} & \textbf{0.88}\\

    \hline
    
    \end{tabular}
    \label{table:lip_generation}
\end{table}

\vspace{-10pt}
\subsection{Model's variation to speaker attributes}

To evaluate the effect of speaker attributes such as gender, language, and accent on speech enhancement, we test our approach on diverse unseen clean speech sources mixed with VGGSound noises. For gender evaluation, we automatically classify the LRS3 test set into male and female speakers using a gender detection tool~\cite{ar2018cvlib}. We also test our model across different languages and accents using clean speech sourced from TTS datasets~\cite{ljspeech,iitmtts}. The results (Table~\ref{table:gender}) clearly demonstrate that there is no distinctive variation in performance across gender of the speakers, but scores do vary across languages and accents, as the training data is mainly comprised of western-accented English.

\begin{table}[ht]
    \centering
    \setlength{\tabcolsep}{3pt}
    \caption{Effect of speaker attributes such as gender, language and accent on model performance.}
    \begin{tabular}{lcccccc}
    \hline
 
    \textbf{Attr.} & \textbf{Class} & \textbf{PESQ} & \textbf{CSIG} & \textbf{CBAK} & \textbf{COVL} & \textbf{STOI}\\
    \hline
    
    \multirow{2}{*}{Gender} & Female & 2.75 & 3.29 & 2.42 & 2.35 & 0.89\\
    & Male & 2.80 & 3.15 & 2.43 & 2.24 & 0.89\\
    
    \hline
    
    \multirow{2}{*}{Lang.} & Hindi & 2.46 & 2.45 & 2.41 & 1.94 & 0.89\\
    & Bengali & 2.39 & 2.86 & 2.45 & 2.14 & 0.86\\
    
    \hline
    
    \multirow{2}{*}{Accent} & Indian & 2.57 & 2.75 & 2.38 & 2.10 & 0.87\\
    & American & 2.60 & 2.89 & 2.49 & 2.16 & 0.89\\
    \hline
    
    \end{tabular}
    \label{table:gender}
\end{table}

\vspace{-10pt}
\section{Applications}
\label{section:applications}
Pseudo-visual speech enhancement opens up multiple application areas that were only being dominated by audio-only methods. The lack of clearly visible real lip movements is very common. For instance, in dynamic scenes like vlogs or press conferences, the camera constantly pans to other elements of the scene. This is even more common in the case of movies, and here, additional downstream applications such as automatic subtitle generation can benefit from speech enhancement. Recently, there is also an increased interest in improving the video call experience in works such as lip to speech synthesis~\cite{Prajwal_2020_CVPR} and talking face generation~\cite{wav2lip:2020}. Psuedo-visual methods are very apt for this application, as it is not possible to expect a high-quality visual stream consistently during video calls due to frequent network slowdowns and interruptions. As lip synthesis methods become more robust and realistic, we expect that pseudo-visual methods can improve further.

\section{Conclusion}
\label{section:conclusion}
In this work, we proposed a novel speech enhancement method that combines the diversity of applications of audio-only works while also exploiting the benefits of lip motion. We do this by generating a pseudo-visual stream for any given noisy input audio. We showed that it is possible to generate accurate and reliable lip motion that reflects the speech component in noisy audios. The generated lip movements serve as a \textit{visual noise filter}, which assists the downstream enhancement model. Consequently, we showed significant gains over traditional audio-only speech enhancement works. Our approach adds a new dimension to the space of face and speech. With suitable modifications, a plethora of other audio-only algorithms like ASR can now be supplemented with additional generated visual information that was not available originally. We believe, exploring various problems with a similar approach could lead to further useful insights into this new space.

{\small
\bibliographystyle{ieee_fullname}
\bibliography{egbib}

\begin{thebibliography}{10}\itemsep=-1pt

\bibitem{Afouras18c}
Triantafyllos Afouras, Joon~Son Chung, Andrew Senior, Oriol Vinyals, and Andrew
  Zisserman.
\newblock Deep audio-visual speech recognition.
\newblock {\em IEEE transactions on pattern analysis and machine intelligence},
  2018.

\bibitem{TheConversation_Afouras_2018}
T. Afouras, J.~S. Chung, and A. Zisserman.
\newblock The conversation: Deep audio-visual speech enhancement.
\newblock In {\em INTERSPEECH}, 2018.

\bibitem{Afouras18d}
T. Afouras, J.~S. Chung, and A. Zisserman.
\newblock Lrs3-ted: a large-scale dataset for visual speech recognition.
\newblock In {\em arXiv preprint arXiv:1809.00496}, 2018.

\bibitem{iitmtts}
Arun Baby, Anju~Leela Thomas, NL Nishanthi, TTS Consortium, et~al.
\newblock Resources for indian languages.
\newblock In {\em Proceedings of Text, Speech and Dialogue}, 2016.

\bibitem{9053174}
H. {Chen}, W. {Xie}, A. {Vedaldi}, and A. {Zisserman}.
\newblock Vggsound: A large-scale audio-visual dataset.
\newblock In {\em ICASSP 2020 - 2020 IEEE International Conference on
  Acoustics, Speech and Signal Processing (ICASSP)}, pages 721--725, 2020.

\bibitem{Dean2010TheQC}
David Dean, Sridha Sridharan, Robbie Vogt, and Michael Mason.
\newblock The qut-noise-timit corpus for the evaluation of voice activity
  detection algorithms.
\newblock In {\em INTERSPEECH}, 2010.

\bibitem{Donahue2018ExploringSE}
Chris Donahue, Bo Li, and Rohit Prabhavalkar.
\newblock Exploring speech enhancement with generative adversarial networks for
  robust speech recognition.
\newblock {\em 2018 IEEE International Conference on Acoustics, Speech and
  Signal Processing (ICASSP)}, pages 5024--5028, 2018.

\bibitem{cocktailparty_erphat_2018}
Ariel Ephrat, Inbar Mosseri, Oran Lang, Tali Dekel, Kevin Wilson, Avinatan
  Hassidim, William~T. Freeman, and Michael Rubinstein.
\newblock Looking to listen at the cocktail party: A speaker-independent
  audio-visual model for speech separation.
\newblock {\em ACM Trans. Graph.}, 37, 2018.

\bibitem{Gao_2018_CVPR_Workshops}
Ruohan Gao, Rogerio~S. Feris, and Kristen Grauman.
\newblock Learning to separate object sounds by watching unlabeled video.
\newblock In {\em The IEEE Conference on Computer Vision and Pattern
  Recognition (CVPR) Workshops}, June 2018.

\bibitem{timit}
J. Garofolo, Lori Lamel, W. Fisher, Jonathan Fiscus, D. Pallett, N. Dahlgren,
  and V. Zue.
\newblock Timit acoustic-phonetic continuous speech corpus.
\newblock {\em Linguistic Data Consortium}, 11 1992.

\bibitem{Germain2019SpeechDW}
François~G. Germain, Qifeng Chen, and Vladlen Koltun.
\newblock Speech denoising with deep feature losses.
\newblock In {\em INTERSPEECH}, 2019.

\bibitem{ljspeech}
Keith Ito et~al.
\newblock The lj speech dataset, 2017.

\bibitem{yst_ijcv_2019}
Amir Jamaludin, Joon~Son Chung, and Andrew Zisserman.
\newblock You said that? : Synthesising talking faces from audio.
\newblock {\em International Journal of Computer Vision}, 2019.

\bibitem{KR:2019:TAF:3343031.3351066}
Prajwal K~R, Rudrabha Mukhopadhyay, Jerin Philip, Abhishek Jha, Vinay
  Namboodiri, and C~V Jawahar.
\newblock Towards automatic face-to-face translation.
\newblock In {\em Proceedings of the 27th ACM International Conference on
  Multimedia}, MM '19, pages 1428--1436. ACM, 2019.

\bibitem{obamanet}
Rithesh Kumar, Jose Sotelo, Kundan Kumar, Alexandre de Br{\'{e}}bisson, and
  Yoshua Bengio.
\newblock Obamanet: Photo-realistic lip-sync from text.
\newblock {\em CoRR}, abs/1801.01442, 2018.

\bibitem{Lu2013SpeechEB}
Xugang Lu, Yu Tsao, Shigeki Matsuda, and Chiori Hori.
\newblock Speech enhancement based on deep denoising autoencoder.
\newblock In {\em INTERSPEECH}, 2013.

\bibitem{ConditionalGA_2017_Michelsanti}
Daniel Michelsanti and Z. Tan.
\newblock Conditional generative adversarial networks for speech enhancement
  and noise-robust speaker verification.
\newblock In {\em INTERSPEECH}, 2017.

\bibitem{Owens_2018_ECCV}
Andrew Owens and Alexei~A. Efros.
\newblock Audio-visual scene analysis with self-supervised multisensory
  features.
\newblock In {\em The European Conference on Computer Vision (ECCV)}, September
  2018.

\bibitem{segan}
Santiago Pascual, Antonio Bonafonte, and Joan Serrà.
\newblock Segan: Speech enhancement generative adversarial network.
\newblock In {\em Proc. Interspeech 2017}, pages 3642--3646, 2017.

\bibitem{Perez_2020_WACV}
Andres Perez, Valentina Sanguineti, Pietro Morerio, and Vittorio Murino.
\newblock Audio-visual model distillation using acoustic images.
\newblock In {\em The IEEE Winter Conference on Applications of Computer Vision
  (WACV)}, March 2020.

\bibitem{lr2_petridis_2017}
Stavros Petridis, Yujiang Wang, Zuwei Li, and Maja Pantic.
\newblock End-to-end audiovisual fusion with lstms.
\newblock {\em arXiv preprint arXiv:1709.04343}, 2017.

\bibitem{lr1_petridis_2017}
Stavros Petridis, Yujiang Wang, Zuwei Li, and Maja Pantic.
\newblock End-to-end multi-view lipreading.
\newblock {\em arXiv preprint arXiv:1709.00443}, 2017.

\bibitem{lr3_petridis_2020}
Stavros Petridis, Yujiang Wang, Pingchuan Ma, Zuwei Li, and Maja Pantic.
\newblock End-to-end visual speech recognition for small-scale datasets.
\newblock {\em Pattern Recognition Letters}, 131:421--427, 2020.

\bibitem{piczak2015dataset}
Karol~J. Piczak.
\newblock {ESC}: {Dataset} for {Environmental Sound Classification}.
\newblock In {\em Proceedings of the 23rd {Annual ACM Conference} on
  {Multimedia}}, pages 1015--1018. {ACM Press}, 2015.

\bibitem{ar2018cvlib}
Arun Ponnusamy.
\newblock cvlib - high level computer vision library for python.
\newblock \url{https://github.com/arunponnusamy/cvlib}, 2018.

\bibitem{Prajwal_2020_CVPR}
K~R Prajwal, Rudrabha Mukhopadhyay, Vinay~P. Namboodiri, and C.V. Jawahar.
\newblock Learning individual speaking styles for accurate lip to speech
  synthesis.
\newblock In {\em The IEEE/CVF Conference on Computer Vision and Pattern
  Recognition (CVPR)}, June 2020.

\bibitem{wav2lip:2020}
K~R Prajwal, Rudrabha Mukhopadhyay, Vinay~P. Namboodiri, and C.V. Jawahar.
\newblock A lip sync expert is all you need for speech to lip generation in the
  wild.
\newblock In {\em Proceedings of the 28th ACM International Conference on
  Multimedia}, MM '20, page 484–492, New York, NY, USA, 2020. Association for
  Computing Machinery.

\bibitem{rix2001perceptual}
Antony~W Rix, John~G Beerends, Michael~P Hollier, and Andries~P Hekstra.
\newblock Perceptual evaluation of speech quality (pesq)-a new method for
  speech quality assessment of telephone networks and codecs.
\newblock In {\em IEEE International Conference on Acoustics, Speech, and
  Signal Processing. Proceedings}, volume~2, pages 749--752, 2001.

\bibitem{wienerfilter}
P. Scalart and J.~V. Filho.
\newblock Speech enhancement based on a priori signal to noise estimation.
\newblock ICASSP ’96, page 629–632. IEEE Computer Society, 1996.

\bibitem{boltzmann_2015_sui}
Chao Sui, Mohammed Bennamoun, and Roberto Togneri.
\newblock Listening with your eyes: Towards a practical visual speech
  recognition system using deep boltzmann machines.
\newblock In {\em Proceedings of the IEEE International Conference on Computer
  Vision (ICCV)}, pages 154--162, 2015.

\bibitem{Suwajanakorn2017SynthesizingOL}
Supasorn Suwajanakorn, Steven~M. Seitz, and Ira Kemelmacher-Shlizerman.
\newblock Synthesizing obama: learning lip sync from audio.
\newblock {\em ACM Trans. Graph.}, 36:95:1--95:13, 2017.

\bibitem{taal2010short}
Cees~H Taal, Richard~C Hendriks, Richard Heusdens, and Jesper Jensen.
\newblock A short-time objective intelligibility measure for time-frequency
  weighted noisy speech.
\newblock In {\em IEEE international conference on acoustics, speech and signal
  processing}, pages 4214--4217. IEEE, 2010.

\bibitem{8461944}
Y. {Tu}, I. {Tashev}, S. {Zarar}, and C. {Lee}.
\newblock A hybrid approach to combining conventional and deep learning
  techniques for single-channel speech enhancement and recognition.
\newblock In {\em 2018 IEEE International Conference on Acoustics, Speech and
  Signal Processing (ICASSP)}, pages 2531--2535, 2018.

\bibitem{ValentiniBotinhao2017NoisySD}
Cassia Valentini-Botinhao.
\newblock Noisy speech database for training speech enhancement algorithms and
  tts models.
\newblock 2017.

\bibitem{ValentiniBotinhao2016InvestigatingRS}
Cassia Valentini-Botinhao, Xiaohua Wang, Shinji Takaki, and Junichi Yamagishi.
\newblock Investigating rnn-based speech enhancement methods for noise-robust
  text-to-speech.
\newblock In {\em SSW}, 2016.

\bibitem{6709856}
C. {Veaux}, J. {Yamagishi}, and S. {King}.
\newblock The voice bank corpus: Design, collection and data analysis of a
  large regional accent speech database.
\newblock In {\em 2013 International Conference Oriental COCOSDA held jointly
  with 2013 Conference on Asian Spoken Language Research and Evaluation
  (O-COCOSDA/CASLRE)}, pages 1--4, 2013.

\bibitem{Wang_2020_WACV}
Jianren Wang, Zhaoyuan Fang, and Hang Zhao.
\newblock Alignnet: A unifying approach to audio-visual alignment.
\newblock In {\em The IEEE Winter Conference on Applications of Computer Vision
  (WACV)}, March 2020.

\end{thebibliography}
}

\end{document}